%% file: main.tex
\documentclass[runningheads]{llncs}
\usepackage[width=122mm,left=12mm,paperwidth=146mm,height=193mm,top=12mm,paperheight=217mm]{geometry}
\usepackage{graphicx}
\usepackage{color}
 \usepackage{graphics} 
\usepackage{times} 
\usepackage{amsmath} 
\usepackage{amssymb}  
 \usepackage{xcolor}
\usepackage{wrapfig}
\usepackage{bm}
\usepackage{array,multirow,graphicx}
\usepackage{amsfonts}
\usepackage[linesnumbered,ruled,vlined]{algorithm2e}
\usepackage{multicol}
\usepackage{scalerel,stackengine}
\newcommand\reallywidehat[1]{%
\savestack{\tmpbox}{\stretchto{%
  \scaleto{%
    \scalerel*[\widthof{\ensuremath{#1}}]{\kern-.6pt\bigwedge\kern-.6pt}%
    {\rule[-\textheight/2]{1ex}{\textheight}}
  }{\textheight}%
}{0.5ex}}%
\stackon[1pt]{#1}{\tmpbox}%
}
\usepackage{hyperref}
\hypersetup{ colorlinks, citecolor=green, filecolor=black, linkcolor=blue, urlcolor=blue }

\graphicspath{{figs/}}

\begin{document}

\title{Deep  execution monitor for robot assistive tasks} 

\titlerunning{Deep execution monitor}

\author{Lorenzo Mauro\inst{1}\orcidID{2-7895-6307} \and Edoardo Alati\inst{1}\orcidID{2-8761-910X} \and Marta Sanzari\inst{1}\and Valsamis Ntouskos \inst{1}\orcidID{3-1810-7802}\and Gianluca Massimiani \and Fiora Pirri \inst{1}\orcidID{1-8665-9807}}
\institute{Alcor Lab, Diag, University of Rome La Sapienza}

\maketitle

\begin{abstract}
We consider a novel approach to  high-level robot task execution for a robot assistive task.  
In this work we explore the problem of learning to predict the next subtask  by introducing a deep model for both  sequencing  goals and for visually evaluating the state of a task.  We show that deep learning for  monitoring robot tasks execution very well supports the interconnection between task-level planning  and robot operations.
These solutions can also cope with the natural non-determinism of the execution monitor.
We show that a deep execution monitor leverages robot performance. We measure the improvement taking into account some robot helping tasks performed at a warehouse. 
\end{abstract}

\section{Introduction}\label{sec:introduction}
\vspace*{-3mm}

In this paper, we present a novel approach to model high-level robot task execution.  
An execution monitor  is a real-time  decision process, which  amounts to choosing at each step of the execution the next subtask and deciding whether the current task succeeded or failed \cite{fikes1971,nilsson1973}. A real-time execution monitor involves plan inference, verification of the current robot state, and choice of next goal state.

Several authors, in the planning community, have explored hierarchical task networks (HTN) (see for instance \cite{erol1994}) and hierarchical goal networks (HGN) (see for example \cite{shivashankar2015})  to provide a way of sequencing  a suitable  decision process \cite{alford2016} at the correct level. However, both HTN and HGN require that these decisions are stacked a priory in the network, putting on the designer the burden to provide a task decomposition, for each task.

In this paper we overcome these difficulties by integrate two deep models to predict next state choice.
The first model is a DCNN,  identifying the objects in the scene and supporting recognition of relations holding at the current execution time. The second  is a sequence to sequence model ({\em seq2seq}) \cite{sutskever2014sequence} with attention \cite{bahdanau2014neural,luong2014,luong2015a} inferring a plausible next robot world-state given the current  world-state.  The interplay between the two models and classical planning grounds the specification of a world-state.
The execution monitor manages the interaction amid the models at execution time.  This is a very preliminary contribution, considering only the high-level robot decisions. Direct robot control is managed by  state charts \cite{wachter2016}.

\paragraph*{\em \bf Main idea and contribution}
In this paper we address a {\em vision-based deep execution monitor} (VDEM) for robot tasks. The main idea is the introduction of a  robot learning model to predict the next goal from the current one, verifying the  preconditions and effects of the currently executed action.  Preconditions and effects are specified in a symbolic language. Whether they hold or not at a state can be determined by the robot vision. The robot  monitors the states of its execution by linking the symbolic language with the vision interpretation, such that the objects in the scene are the terms  of the symbolic language, and the relations are the predicates. 
\begin{figure}[t!]
\centering
 \includegraphics[width=0.95\columnwidth]{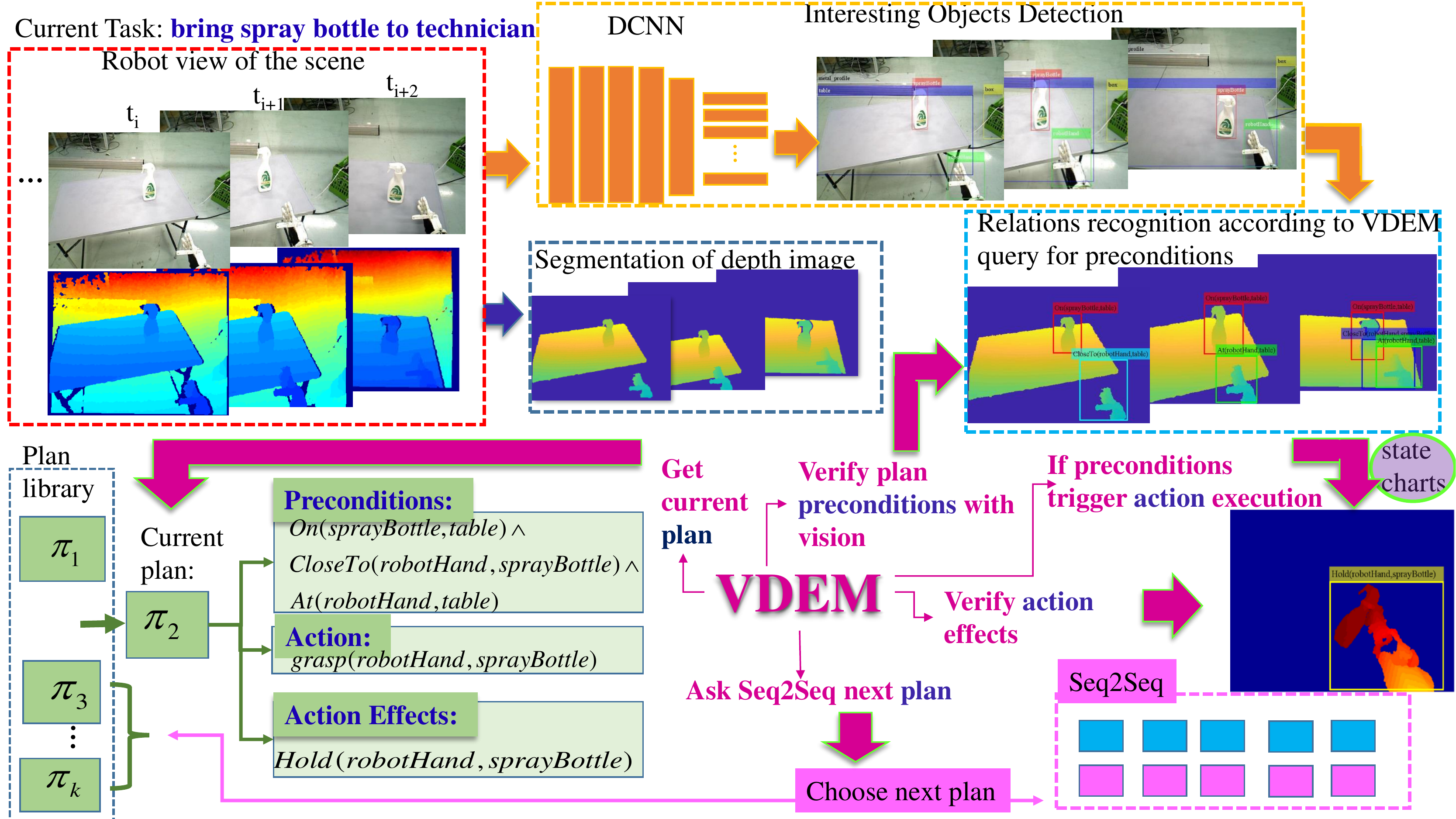}
\caption{The schema above presents the flow of information managed by the  deep execution monitor (VDEM) for the  task {\em bring the spray bottle to the technician}. While the robot observes the scene, the state is built by the detected relations, restricted according to what is  required by the current planner. The VDEM queries the vision system to both verify the preconditions for action to be executed,  and the realization of the action effects. A plan library (see e.g. \cite{hofmann2017} for a reference) provides background knowledge  in a symbolic language. The seq2seq model learns to predict goal-states, according to the specific task and current state, and it is invoked by  the VDEM whenever a new goal state is required. }\label{fig:schema}
\end{figure}
 The next goal state is inferred by associating to each goal described by some plan in the plan library, the goal which is the most plausible successor state. Therefore, given that $X$ is a goal descriptions, and $Y$ is the next goal description, the seq2seq model infers $P(Y|X)$.  A description is formed by the predicates and terms verified by vision, which form the current robot world-state. The seq2seq model is formed by an encoder fed by token of the symbolic language, an attention mechanism that pairs each description with the task, which is a sort of memory of the goals concerned with such a task, and a decoder, which infer the most likely successor state.

Though recent approaches   \cite{lenz2015,Al-Omari-2016,ZhuGKFFGMF17,mirowski2016,zhu2017} have considered vision based execution, our approach is novel in combining vision based execution with next step prediction,  binding the planning symbolic languages with visual instances. The binding allows the execution monitor to generate a  state merging  vision and planning feedbacks. Furthermore, the approach provides both depth and location for relations recognition  to  cope with the task dynamics.  

 We tested the framework at a warehouse with a humanoid robot, described in the experiment section, see Section \ref{sec:experiments}. We provide ablation of the execution monitor functionalities  to experiment the robotic performance and the advantages of the proposed vision based deep execution monitor.

\typeout{-------------------- RELATED -----------------------}
\section{Related Work}\label{sec:relwork}
\paragraph{\em \bf Vision based robot execution}
The earliest definitions of execution monitoring in nondeterministic environments were introduced by \cite{fikes1971,nilsson1973}. Since then an extraordinary amount of research has been done  to address  the nondeterministic response  of the environment  to robot actions. Several definitions of execution monitoring are reported in \cite{pettersson2005}. For high-level robot tasks, a review of these efforts is given in \cite{ingrand2017}. The role of perception in execution monitoring was already foreseen in the work of \cite{doyle1986}. Likewise, recovery from  errors that could occur at execution time was already faced by \cite{wilkins1985}. 
Despite this foresight, the difficulties in dealing with scene recognition have directed the effort toward models managing the effects of actions  such as  \cite{sutton1998,bertsekas1995},  allowing to execute actions in partially observable environments, similarly as in \cite{boutilier2000},\cite{gianni2015,finzi2001}.   On the other hand, different approaches have studied learning policies for planning as in \cite{littman2002} and also for decision making, in partially observable domains \cite{haarnoja2016}. Vision based planning has been studied in \cite{ZhuGKFFGMF17}. These approaches did not consider execution monitor and the duality between perception and learning. Likewise despite facing the integration of observations in high-level monitor   \cite{hornung2014,mendoza2015} did not use perception for verifying the current state, which is crucial for both monitoring and further decision learning.   

\paragraph{\em\bf Relations recognition in videos}
Relations in videos  dynamically change, in the sense that the configuration of the involved objects is altered according to the robot vantage points. Recently a number of approaches have studied spatial relations and their grounding, such as \cite{guadarrama2013,Lu-2016,das2016,santoro2017}. Among them, only \cite{guadarrama2013} faces the problem from the point of view of robot task execution. There are also recent contributions  concerned with human activity recognition and human-objects interaction studying  the problem regarding human dynamics such as  \cite{ntouskos2015,sanzari2016,wu2017,zhu2017},  and  \cite{wang2017}, here in particular for container and containee relations. Although these latter approaches consider both videos and 3D objects they do not face general relations amid objects. The main difficulty seems that of recognizing relations in a complex scenario without overloading the perceptual scene, namely what the robot has to infer from the scene.  To this end, and also to maintain real time execution,  we rely on   the execution monitor querying the visual interpretation at each current state about the existence of specific relations. Relations computation exploits  approximate depth estimation within the object bounding box.  To obtain this good performance we use  DCNN trained on different classes of models, which are retrieved by the execution monitor, and  the active features of the recognized objects, involved in the relation, to estimate the object depth. 

\paragraph{\em \bf Sequence to sequence models and next step prediction}
Sequence to sequence models (seq2seq)\cite{sutskever2014} are made of two networks, one  for processing the input and a second network generating the output, in an encoder-decoder configuration. They have  shown an excellent performance in several sequence prediction problems especially in  machine
translation,  image captioning  and even  in high-level decision processing.
In planning problems, \cite{karkus2017} have proposed recently QMDP-Net combining POMDP and LSTM to obtain a neural network architecture under partial observability. They applied their model to 2D grids to cope  with 2D path planning. While we do not know of other approaches to execution monitor and high-level planning with seq2seq architecture, LSTM have been used for path planning, while \cite{gupta2017} show that their CMP approach to navigation outperforms LSTM. The introduction of an attention mechanism
 \cite{bahdanau2014neural,luong2014,luong2015a} has improved sequence to sequence models essentially for neural translation and also for image captioning. Attention mechanisms for robot execution have been studied in \cite{ntouskos2013}, and here in particular we base our approach on the attention mechanism to exploit the task context. 

The problem of predicting next step has not yet faced with seq2seq models. An approach to driving the focus of attention to the next useful object has been introduced by \cite{furnari2017next}. On the other hand \cite{damen2018} have designed a new public database including annotations also for the next action, which is relevant for execution monitor, where prediction of next state can take advantage of surrounding people actions.

\typeout{--------------------  OVERVIEW  -----------------------}
\section{ Deep execution monitoring}\label{sec:overview}
In this section we give an overview of the execution monitor (VDEM) altogether, providing at the end  of the section the main algorithms. 

\noindent
\paragraph{\em \bf Preliminaries on the environment and the tasks}
We consider robot assistive tasks related to maintenance activities  at a warehouse. The robot language ${\mathcal L}$ is defined by atoms, which are formed by predicates taking terms as arguments. Terms, can be either variables or constants, and they are instantiated by the objects that the robot identify in the environment. Likewise, predicates are the relations the robot is able to identify in the environment. Predicates take  also indexed terms denoting the frame as arguments. The robot language ${\mathcal L}$ is extended with meta terms denoting task, hence ${\mathcal L}\cup\{T\}_{i=1}^K$.  Where $T_i$ is a sentence specifying a task. Tasks sentences are, for example, {\em pass  the brush and the cloth to the technician}, {\em  help the technician to hold the guard}. 
Therefore a task sentence is expressed in natural language, and the execution of a task requires a number of actions to be performed, for both controlling the robot visual process and the robot motion. These actions are specified by plans collected into the plan library. 

\begin{equation}
\begin{array}{l}
VisionOn(robot,t_0) \wedge Free(robot\_hand,t_0), \\
Detected(brush,t_1)\wedge Detected(ladder,t_1)\wedge
On(brush,ladder,t_1),\\ 
At(robot,ladder,t_2) \wedge  Holding(robot\_hand,brush,t_3), \\
 Detected(technician,t_3)\wedge CloseTo(robot,technician,t_3),  \\
Detected(technician\_hand,t_4) \wedge Holding(technician\_hand,brush,t_4) \wedge\\
 Free(robot\_hand,t_{4}) 
\end{array}
\end{equation}

\paragraph{\em\bf Plans and plan library} Let us assume that the execution of a task requires the execution of $n$ plans, where each plan specifies a number of actions.

{\em A plan library is a collection of actions}. 
In a plan library, each plan defines all the actions needed to achieve a goal of a  part of a  task,  by a suitable axiomatization. For example, to grasp an object the robot needs to be close to the object, which is a partial task.

A plan is formed by a {\em problem} specifying the initial state and a goal, defined in the propositional Planning Domain Definition Language ({\em pddl}),  and by a {\em domain} providing an  axiomatization of actions, which is first-order {\em pddl} with types and equality.  Plans, therefore, form the background knowledge of the robot about what is needed for an action  to be performed.

  A state $s$,  with respect to an action $a$, is formed by either the preconditions for executing $a$ or by the
effects of $a$ execution. When  $s$ is a goal state this is  the goal of the {\em problem}.  To  simplify the presentation here we assume that the preconditions and effects are  conjunctions of
binary or unary atoms, and a state can be reduced to $s = \bigwedge_i R_i(\nu_{i1},{\ldots},\nu_{ik},t)$, where $(\nu_{i1},{\ldots},\nu_{ik})$, $k>=1$, are ground terms. Plan inference amounts to deduce the goal of the problem, given the starting state.  A goal of a problem is, for example, $At(robotHand,table)$, requiring to search where the table is, and reaching it.

To facilitate  inference, the set of actions axiomatized in a plan domain are partitioned into  actions that  affect  the state of the world (like moving objects around) and  ecological actions, which affect only the state of the robot.  Ecological actions are for example  $search$, $verify\_vision$, $turn\_ head$, $look\_up$, $look\_down$.
A plan is formed by at most a single action that can affect  the  world  and by a number of ecological actions. This allows to partition  the terms of the plan  into  terms denoting the world, with their types hierarchy, and terms related to the robot representation, requiring appropriate measures, for vision and motion control.

The plan library is the collection of all plans needed for the assistive tasks and it is generated together with the maintenance experts to cope with the foreseen assistive tasks, hence the hypothesis is that: 
{\em for all foreseen tasks there exists a sequence of goals factoring them.}

\paragraph{\em\bf Task factorization} 

Given  a task, factorization amounts to decompose the task into plans, which are supposed to belong to the plan library. 
Task factorization is crucial for a number of issues.  It avoids useless combinations of unrelated groups of objects, it limits the inference of a goal just to the involved objects, it ensures a high flexibility in robot execution, and allows to easily recover from failures.  A top down factorization, such as  HTN  \cite{erol1994} or HGN \cite{shivashankar2015}, might  be too costly to be achieved in real-time, and also might not be able to take care of the state resulting  after the execution of the n-th plan. An incongruence would require, in fact, to search backward for a previous reliable state.

The solution we propose here is to  learn to predict  the next goal, given the current goal state. In this way, given a task and its initial state goal, a successor state goal can be predicted after the success of the current goal state is confirmed.

\input{ALGO}

\noindent
\paragraph{\em\bf Execution}
 The execution monitor loops over the following operations: 1) get the next goal; 2) identify the plan for the given goal;  3) forward the inferred actions  to the {\em state charts}  \cite{wachter2016},  as soon as the preconditions are satisfied, according to the vision process;  4)  verify the effect of the inferred actions; 5) if the current plan goal is obtained ask the seq2seq model to infer next goal and go to 2)  else continue with the current plan. 
The execution, illustrated in Figure \ref{fig:schema}, is resumed in Algorithms \ref{algo:vbdem},\ref{algo:lstm},\ref{algo:vision}. 

Note, therefore, that according to the algorithms the  seq2seq model is called only if the current state is either a  goal of the current plan, just concluded, or the start state of a task. Note that in case of failure a new task ${\mathcal T}'$ can be recovered from last successful state.

\typeout{--------------------  The VISUAL STREAM -----------------------}
\section{Vision Interpretation}\label{sec:visualstream}
\begin{figure}[t!]
\centering
   \includegraphics[width=0.95\columnwidth]{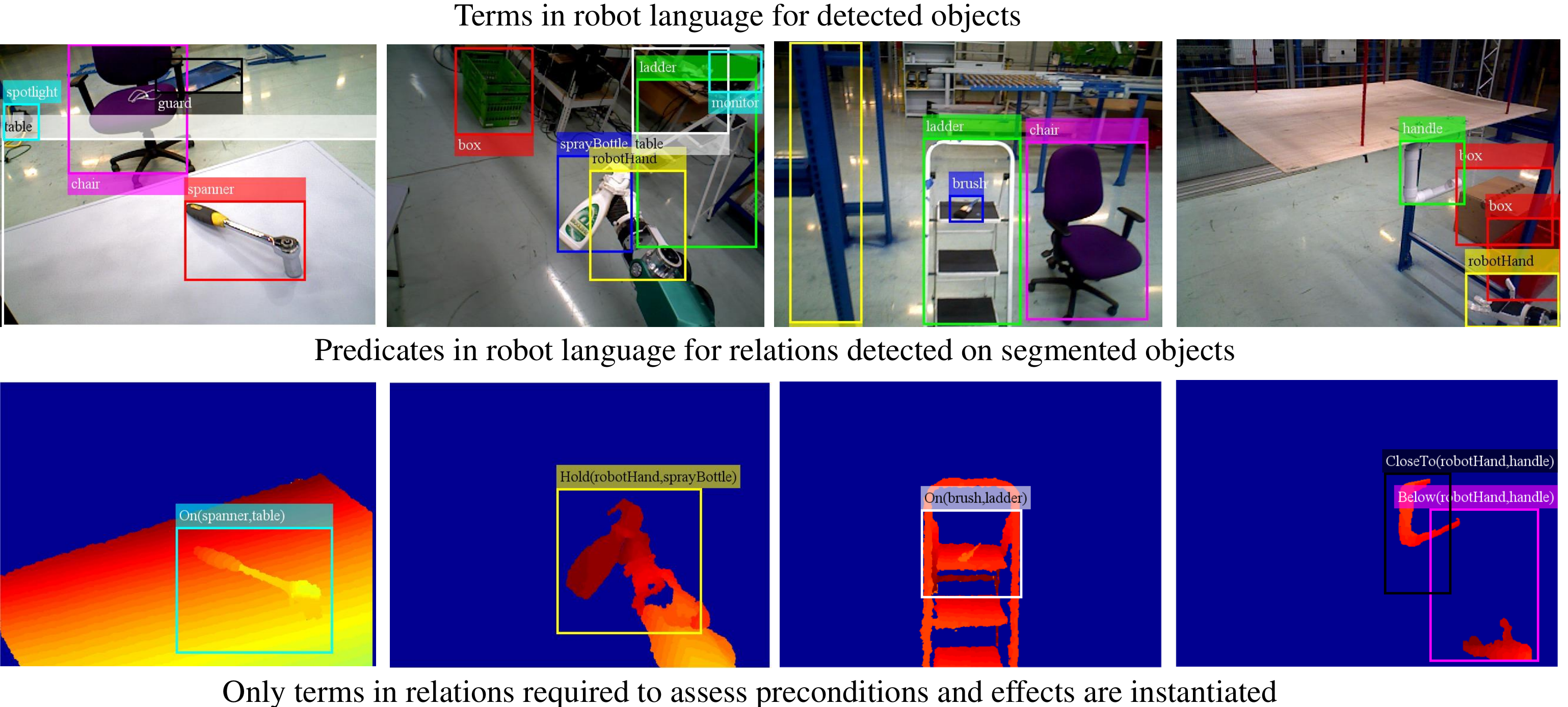}   
\caption[]{Objects detected in the scene observed by the robot, while it is executing its task, are terms of the robot language. Only relations needed by the planner and queried by the VDEM to vision are  considered and instantiated with detected terms.}
   \label{fig:detection}
\end{figure}
As highlighted in the previous section, the execution monitor gets from the current plan the state to be verified in the form of a conjunction of atoms, and query the {\em vision interpretation} to assess if the current state holds. An example is shown in Figure \ref{fig:detection}. 

To detect both objects and relations we have trained Faster R-CNN \cite{ren-2015} on ImageNet \cite{krizhevsky2012}, Pascal-VOC \cite{Everingham15} dataset, and with images taken on site. 
We have trained 5 models to increase accuracy, obtaining a detection accuracy above $0.8$. The good accuracy is also due to a confidence value measured on a batch of 10 images, taken at $30fps$, simply computing the most common value in the batch and returning the sampling mean accuracy for that object.

The model is called according to the state request. For example, if $On(brush,ladder)$ is requested from the plan state, the execution monitor asks the vision interpretation to call the models for $brush$ and $ladder$ first and for $On$ relation for all the found terms, after. Though the main difficult part is searching the objects and the relations, we shall not discuss this here. 

To infer spatial-relations we have introduced a look-up table for the definition of each relation of interest for the assistive task. The relations require the depth  within the bounding boxes of each object denoted by the queried terms.  Depth is crucial in the warehouse environment, because objects at different distances appear within the bounding box of an object, as shown in the first image of Figure \ref{fig:detection}. There is, indeed a tradeoff between using MaskRCNN \cite{he2017} and Faster. With Mask we have the depth segmentation immediately, by projecting the mask on the RGBD image, but objects of the warehouse need to be manually segmented. On the other hand Faster using Imagenet offers a huge amount of data, but depth needs to be obtained. In this version of our work we considered Faster  R-CNN \cite{ren-2015} and did a local segmentation by clustering.

We have first trained a non-parametric Bayes model to determine for each object of interest the number of feature classes. To this end,  we estimate a statistics of the   active features  with dimension $38\times 50 \times 512$, taken  before the last pooling layer, at each pixel inside the recognized object bounding box  (here we are referring to VGG, though we have considered also ZF, see \cite{zeiler2014,simonyan2014}).  Once the number of classes for each object is established we have trained a normal mixture model on the selected feature classes for each object, resulting in a probability map that a pixel belongs to the specific class of the object. 

During execution, as the object is known, we choose the learned parameters for the model to estimate a probability that the pixel in the bounding box belongs to the object.  The distribution on the bounding box is  projected onto the depth map and  a ball-tree  is built using only the pixels with a probability greater than a threshold (we used 0.7). Using unsupervised nearest neighbor, checking the distance, a resulting segmentation is sufficiently accurate for the task at hand. Depth is considered relative to the robot-camera. See Figure \ref{fig:detection}.

Having the depth, the relations are established, a reference are the spatial relations based on the connection calculus \cite{cohn2001qualitative}, though here distance and depth play a primary role, which are not considered in \cite{cohn2001qualitative}.  
To establish the relation amid $n\leq 3$  objects we consider the distance first (within a moving visual cone with vertex the center of projection)  and  further the other properties consistently with the connection calculus and its 3D extensions (see \cite{sabharwal2011}).
See Section \ref{sec:experiments} for an overview of the relations and the accuracy on the recognition. 

\section{The seq2seq architecture for deep monitoring}\label{sec:lstm}

As gathered in previous sections the robot is given a {\em high level task} specified by a sentence, such as {\em help the technician to support the guard}. The objective, here,  is to find the sequence of plans, in the plan library,  ensuring the task to succeed. We have seen that relevant steps to this end are the definition of states, which are conjunctions of literals, inferred by the plans and verified by the vision interpretation to hold before or after the robot executes an action.

We have also introduced the notion of {\em goal state} as the state of a plan problem in which the goal holds. When a goal state is achieved, task execution requires to predict  the next goal, in so ensuring to progress in the accomplishment of the assigned task. 

We show that a sequence to sequence architecture is effective for mapping a current goal state, expressed as a conjunction of literals into a new goal state, where it is intended (see Section \ref{sec:overview}) that the predicted goal is a goal of some plan in the plan library. 

A sequence to sequence system mapping a state of the robot into a new state is a network modeling the conditional probability $p(Y|X)$ of mapping a source state $x_1,\ldots, x_n$ into a target state $y_1,\ldots,y_m$. The encoder-decoder is made of two elements: an encoder which transform the source into a representation $S$ and a decoder generating one target item at a time, so that the conditional probability is  \cite{luong2015a}:
\begin{equation}
\log p(y|x) =\sum_{j=1}^m \log p(y_j | y_1,\ldots y_{j-1},S)
\end{equation}

\typeout{Robot language}

We define an input state as a set of tokens belonging to the extended robot language ${\mathcal L}\cup\{T\}_{i=1}^K$ with ${\mathcal L}$ the language including terms (denoting objects in the scene) and predicates, denoting relations in the scene, and with $T_i$ a task sentence.
Given an input state ${\bf s} = (u_1,\ldots, u_n)$, this is initially mapped into a low dimensional vector ${\bf x}$. With ${\bf x} = W{\bf s}$, where $W$ is the embedding matrix, which is fine-tuned during the training of the seq2seq model.

Given the encoded sequence ${\bf x}$ and the true output sequence ${\bf y}$, encoded as well, the goal is to learn how they match in order to predict, at inference time, the correct ${\bf y}'$ given the input ${\bf s}'$. 

Attention \cite{bahdanau2014neural}, \cite{raffel2017} has become, recently a hot topic for measuring similarities and dissimilarities between input and output sequences, according to the specific objective of the mapping. For example, while in neural machine translation (NMT) alignment can be quite relevant, in the case of a new state prediction alignment is not really relevant while the task at hand it is, since a new goal is looked for while a specific task is executed. 
In general attention computes the relevance of each token in the encoded sequence with respect to the true encoded sequence ${\bf y}$ via a function $\varphi(x_i,{\bf y})$, which returns a score whose distribution, via a softmax function, determines the relevance of each token in ${\bf x}$ with respect to the encoded output ${\bf y}$. This can be expressed as the expectation of a token given the distribution induced by the score:
\begin{equation}
\sum p(z = i | {\bf x}, {\bf y}) x_i
\end{equation} 
Where $p(z = i | {\bf x}, {\bf y})$ is the distribution induced by the softmax applied to the score given to each token $x_i$, with $z$ the indicator of the encoded input tokens. In the literature different score function have been proposed, e.g. additive or multiplicative \cite{bahdanau2014neural,luong2014}:
\begin{equation}
\begin{array}{ll}
\varphi(x_i,{\bf y})= w^{\top} \sigma(W^{(1)} x_i +W^{(2)}{\bf y}) & \mbox{(additive)}\\
\varphi(x_i,{\bf y})=  \langle W^{(1)} x_i +W^{(2)}{\bf y}) \rangle & \mbox{(multiplicative)}
\end{array}
\end{equation}

\noindent Where $W^{(i)}$ are learned weights. 
In our case we have two basic structures, the task sentence and the sequence of atoms. We have also  specific separators: for the atoms $\langle eoa\rangle$, for the end of task sequence $\langle ets\rangle$ and for the end of the state description $\langle eos\rangle$. The attention mechanism required here needs to score the compatibility of  each atom, namely a subsequence of the output sequence ${\bf y}$,  with the task and  with each input token. For example we expect that in the context of the task {\em pass the brush to technician} the output subsequence {\em Hold, technician, brush} has an encoding similar to {\em Hold, robot, brush}  while this is not true in the context of the task {\em help the technician to hold the guard}, in which the correct subsequence would be {\em On, table, brush}. 

To this end we formulate the input and output embedded sequences in terms of subsequences ${\boldsymbol \tau}^{\bf x}= (\tau_1^{\bf x},\ldots \tau_K^{\bf x})$ and ${\boldsymbol \tau}^{\bf y} =(\tau_1^{\bf y},\ldots, \tau_m^{\bf y})$, using both the $\langle eoa\rangle$ and $\langle ets\rangle$, in order to compute the weights of the attention mechanism.   Weights are learned by a dense layer taking as input the concatenation of the previous predicted $\tau^{\bf y}_{t-1}$, from the decoder, the embedded task, which is always $\tau_1^{\bf y}$, and the previous hidden state of the decoder.  The weights for each $\tau$ form a matrix, hence we obtain:  
\begin{equation}
\varphi(\tau_i^{\bf x},{\boldsymbol \tau}^{\bf y})= W^{\top} \sigma(W^{(1)} \tau_i^{\bf x} +W^{(2)}{\boldsymbol \tau}^{\bf y}) 
\end{equation}
Finally, following the softmax application, we have a prediction of the importance of each token  of the encoder according to the 'context' atom and according to the task. Thus we have $p({\bf z} = i|\tau_i^{\bf x}, x_i,{\boldsymbol \tau}^{\bf y})$, which is a vector of the dimension of $\tau_i^{\bf x}$.  This is  the probability that a subsequence, namely an atom, is relevant for the current task and the predicted sequence. Then the output is obtained as the expectation over all the atoms:
\begin{equation}\label{eq:out}
s= \sum p({\bf z} = i|\tau_i^{\bf x}, x_i,{\boldsymbol \tau}^{\bf y})\tau_i^{\bf x}
\end{equation}

We can note that in (\ref{eq:out}) also words are made pivotal, since the probability is a vector. 
 For example, in case the task is {\em bring the brush to the technician}, the $brush$ is a pivotal word, and the context will most probably imply that the mapping of the predicate $Hold$ is from $Hold(robotHand,brush)$ to $Hold(technician,brush)$ and the task sentence triggers attention to both the term $brush$ and the relation $Hold$.

\paragraph{\em\bf Data Collection for the seq2seq model}
The robot vocabulary is formed by 18 unary predicates, 13 binary predicates and 42 terms.  We build the Herbrand Universe from predicates and terms, obtaining a language of more than 35k atoms. Elements of the language are illustrated in Figure \ref{fig:multGoalProp}. 

A number of the atoms does not respect the type hierarchy, which is defined in {\em pddl}, therefore are deleted from the language. Finally we have grown all the goal states provided in the plan library up to 20k states. 

Some of the predicates from the whole set are listed in Table 1, detailing the recognition ability of the vision interpretation. We should note that a number of predicates concerns the robot inner state, such as for example {\em VisionOn} or the head and body positions, which are not listed in Table 1.

\typeout{-------------------- EXPERIMENTS -----------------------}

\section{Experiments and results}\label{sec:experiments}
\paragraph{\em\bf Experiments setup}
Experiments  have been done at a customer fulfillment center warehouse, under different conditions in order to test different aspects of the model. To begin with, all experiments have been performed with a humanoid robot, created at the High Performance Humanoid Technologies Lab (H$^2$T).
The robot has two 8-DoF torque-controlled arms, two 6-DoF wrists, two underactuated 5-finger hands, a holonomic mobile base and 2-DoF head with two stereo camera systems and an RGB-D sensor. 
The Asus Xtion PRO live RGB-D camera has been mounted on the robot head  to provide the video stream to the visual system and ran the \textit{VDEM} on two of the  computers mounted on the robot. We dedicated one to the planning and management of the execution and another one, equipped with an Nvidia Titan GPU, to ran the \textit{visual stream}. Robot control is interfaced with the VDEM via the state charts \cite{wachter2016}.

\paragraph{\em\bf Results for the visual stream}
We trained the visual stream system using images taken from the ImageNet dataset, Pascal VOC, as well as images collected inside the warehouse by the RGB-D camera of the robot. Most of the objects, indeed, are specific of the warehouse and cannot be found in public databases.  The relations considered were essentially those relevant to the  maintenance tasks (see Table \ref{tab:ablat}). To train the DCNN models we split the set of images in training and validation sets with a proportion of 80\%-20\%. We trained a number of  different models for the different types of objects, and we performed 70000 training iterations for each model on a PC equipped with 2 GPUs.
The visual stream has been tested under different conditions, in a standalone tests and during the execution of different tasks. The accuracy has been computed considering the batch of 10 images, accuracy of objects recognition and relations recognition is shown in Table \ref{tab:ablat},  evaluating accuracy and ablation study specifically for relations. 

Mean average precision mAP for object detection is 0.87 and localization in depth is 0.98 accurate up to 3 meters.

\begin{table}
\centering
\resizebox{0.65\columnwidth}{!}{
\begin{tabular}{|l|c|c|c|c|c|c|}
\hline
\textbf{Predicate} & \textbf{Full} & \textbf{BB} & \textbf{ masks} & \textbf{no prior} & \textbf{no shape} & \textbf{no depth} \\[2mm]
CloseTo  & 89\% & 79\% & 82\% & 79\% & 72\% & 49\% \\
Found  & 95\% & 85\% & 81\% & 85\% & 80\% & 61\% \\
Free & 91\% & 86\% & 91\% & 86\% & 83\% & 68\% \\
Hold & 88\% & 72\% & 82\% & 75\% & 74\% & 56\% \\
Inside & 87\% & 64\% & 78\% & 71\% & 65\% & 57\% \\
On & 96\% & 77\% & 85\% & 79\% & 78\% & 65\% \\
InFront & 95\% & 81\% & 85\% & 84\% & 83\% & 63\% \\
Left & 95\% & 81\% & 88\% & 85\% & 86\% & 72\% \\
Right & 91\% & 79\% & 88\% & 79\% & 80\% & 61\% \\
Under & 88\% & 76\% & 69\% & 79\% & 76\% & 59\% \\
Behind & 81\% & 78\% & 78\% & 76\% & 79\% & 61\% \\
Clear & 82\% & 75\% & 80\% & 73\% & 73\% & 60\% \\
Empty & 83\% & 72\% & 78\% & 79\% & 68\% & 61\% \\
\hline
\textbf{Average} & \textbf{89\%} & \textbf{77\%} & \textbf{83\%} & \textbf{79\%} & \textbf{77\%} & \textbf{62\%}\\
\hline
\end{tabular}}
\caption{Accuracy and ablation study of predicate grounding. \textbf{Legend:}  \textit{BB}: bounding boxes only, \textit{masks}:  segmentations masks only, \textit{no prior}: without use of distance \textit{no shape}: without use of shape properties \textit{no depth}: without use of depth. }\label{tab:ablat} 
\end{table}

\paragraph{\em\bf Results of the seq2seq}
\begin{figure}[ht!]
 \centering
  \includegraphics[width=0.48\columnwidth]{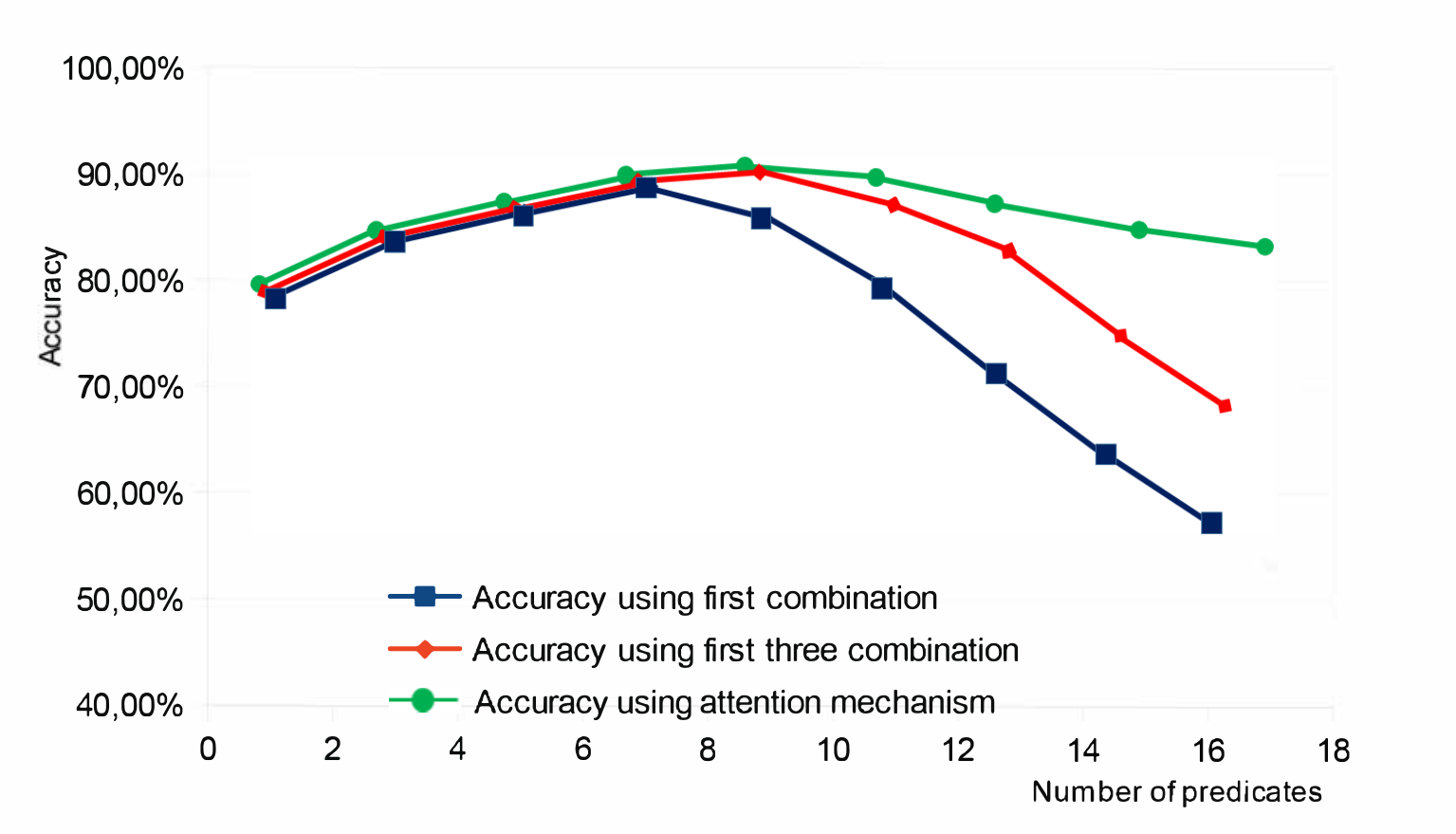},
   \includegraphics[width=5cm, height=4cm]{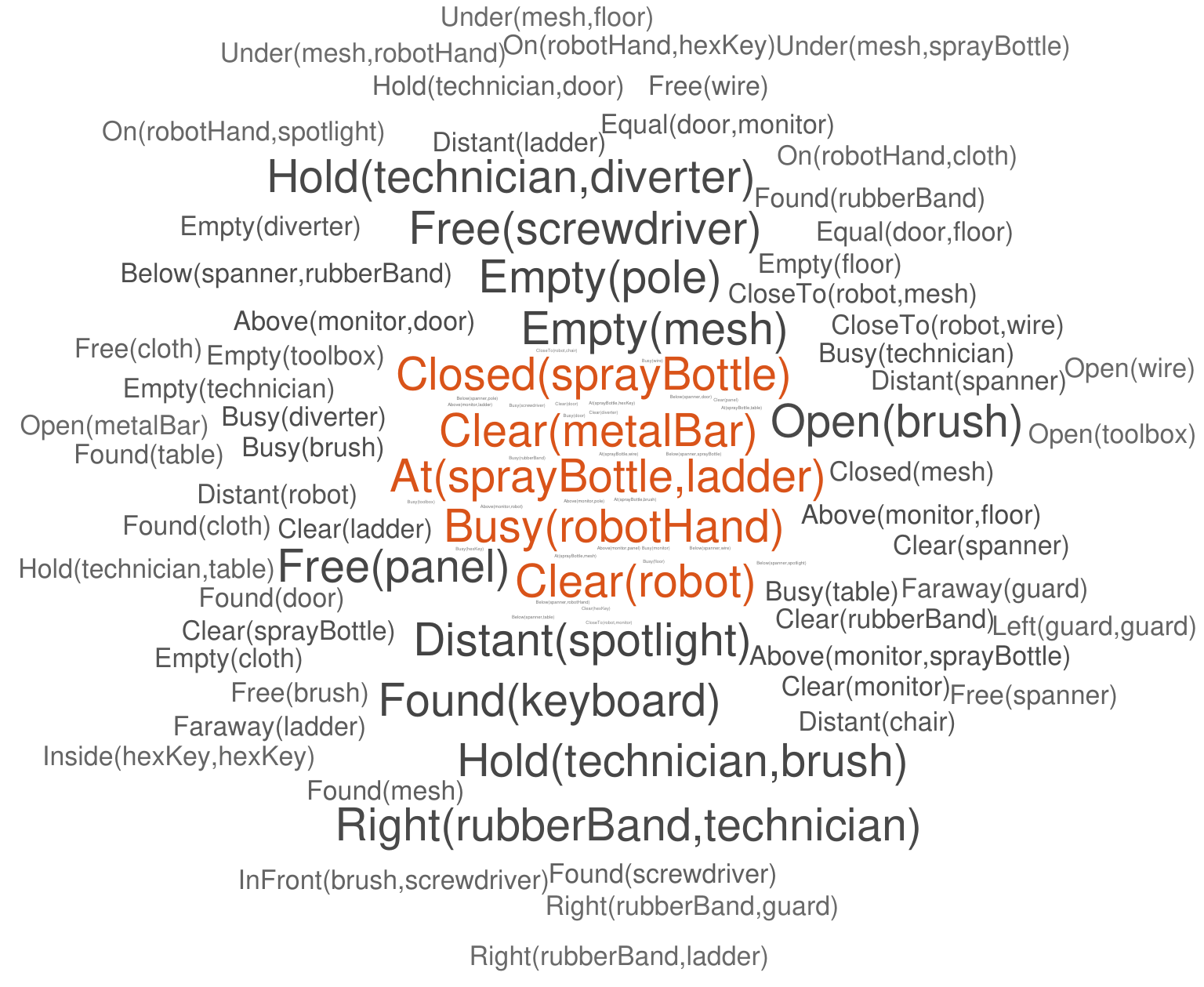}
 \caption{Accuracy plot at variable number of predicates in the input sequence, considering the first combination, the first three and, finally with attention. On the right a cloud representation of the robot language expressed in the form of an Herbrand Universe, namely, all predicates are instantiated with all terms.}\label{fig:multGoalProp}
\end{figure}
We used for the seq2seq network the encoder decoder structure with LSTM \cite{hochreiter1997}, in particular a multilayer bidirectional LSTM for  the encoder. The maximum input sentence length is set to 17 predicates and a task, which is equivalent to 72 words among relations and terms.  The embedding layer transforms the index encoding of every word in the input into a vector of size 20, the encoder then uses a bidirectional LSTM and an LSTM to transform the input question in a vector of size 10.
This vector is repeated 3 times, as the length of output sentence and then it is fed to the decoder network. A fully connected layer is then applied to every time sequence returned and then it is passed to a softmax activation layer.
The attention function is modeled by a fully connected two layers network. 

The seq2seq training uses the Categorical Cross Entropy loss and Adam as an optimizer using batches of 5 sequences for a total of 100 epochs. The total size of the dataset is of 20 thousand sequence pairs.

The accuracy, calculated as the percentage of correct prediction made on a test set extracted from the dataset is used to evaluate the training results. The measurement is done under three different hypotheses.
First we considered only the best combination, then we considered the first three combinations, randomly changing the length of the input sequences, finally we considered the accuracy under the local attention model. 
As shown in Figure \ref{fig:multGoalProp} we vary the number of predicates from one to nineteen, which is equivalent to a sequence length varying from 4 to 72 considering both relations and terms.
It is possible to see that initially the accuracy increases as the amount of atoms increases, this is caused by the fact that with more than one atom the sequence is more specific and characteristic. The maximum accuracy is reached at seven atoms with 94,2\% of accuracy for the first combination and 97,9\% using the first three. After this point the accuracy starts to decrease with the increase of the atoms in the input sequence. On the other hand we can note that by adding the attention mechanism the accuracy keeps high also with a large number of atoms.

 \begin{figure}[t!]
    \includegraphics[width=.33 \textwidth]{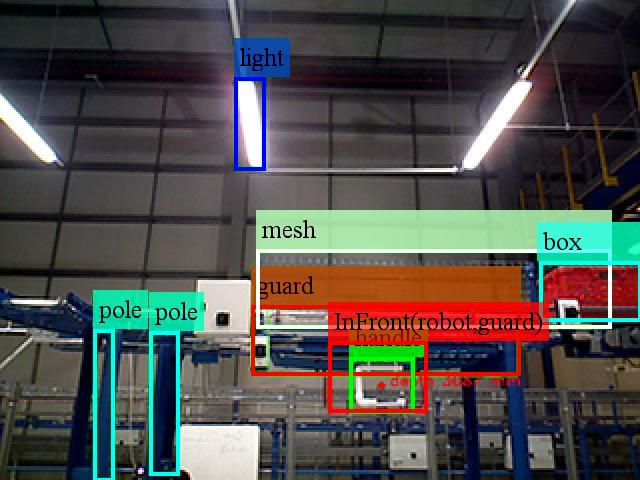}\hfill
    \includegraphics[width=.33 \textwidth]{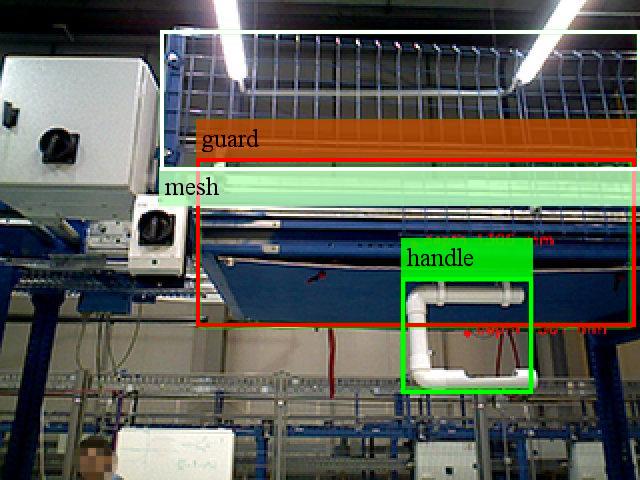}\hfill
    \includegraphics[width=.33 \textwidth]{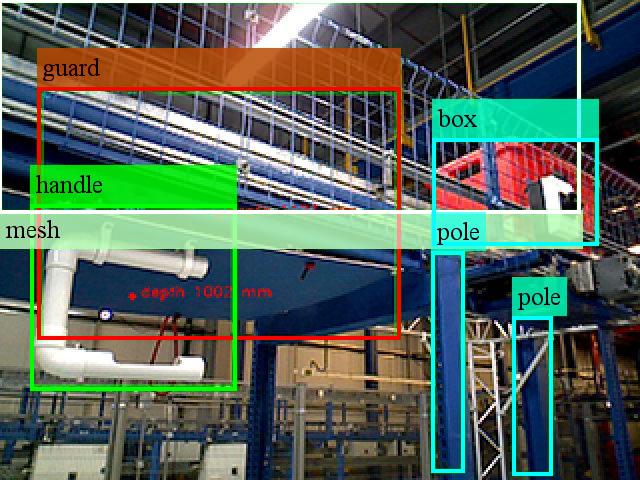}\\
    \includegraphics[width=.33 \textwidth]{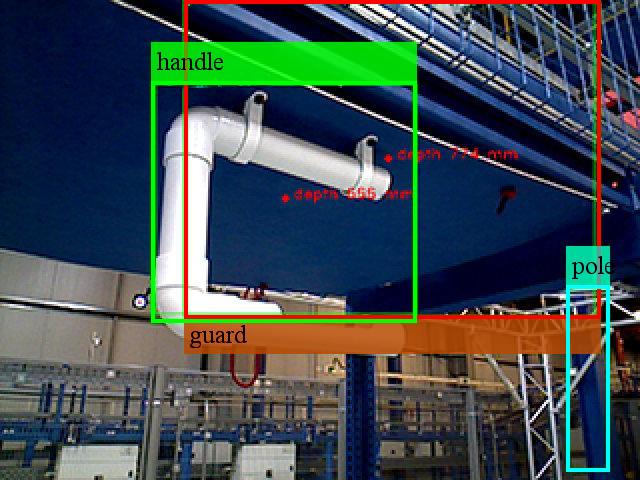}\hfill
    \includegraphics[width=.33 \textwidth]{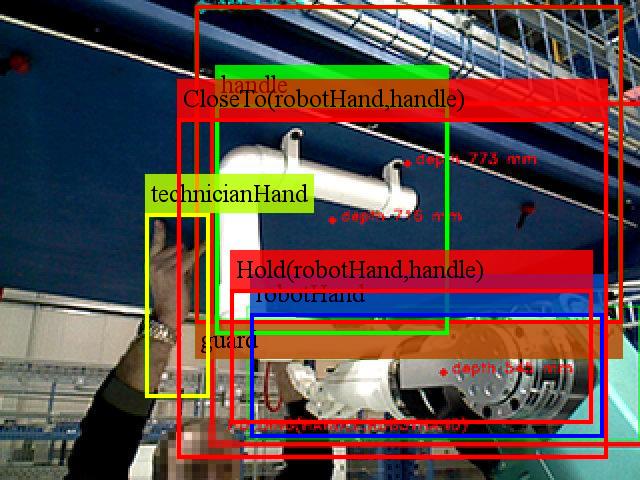}\hfill
    \includegraphics[width=.33 \textwidth]{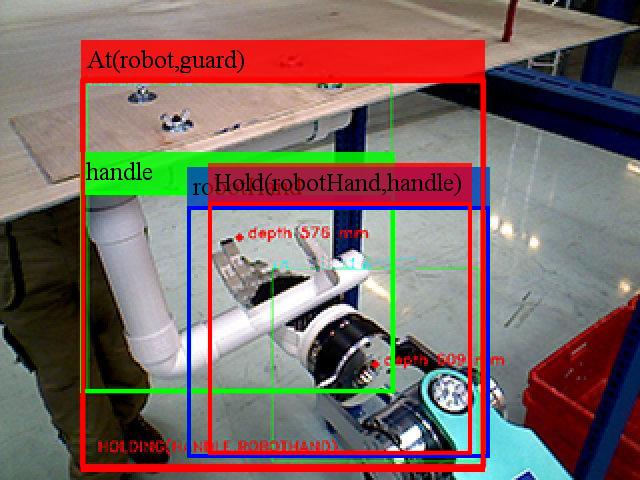}\hfill\\
    \includegraphics[width=.33 \textwidth]{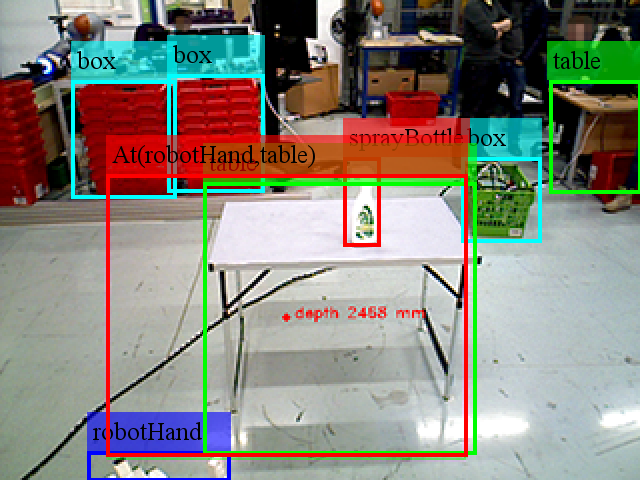}\hfill
    \includegraphics[width=.33 \textwidth]{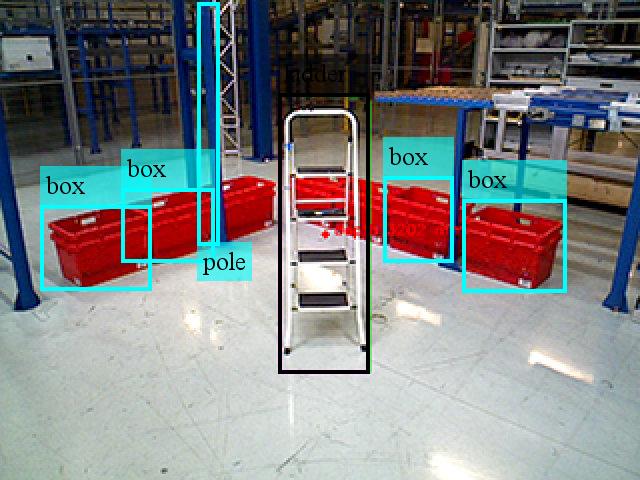}\hfill
    \includegraphics[width=.33 \textwidth]{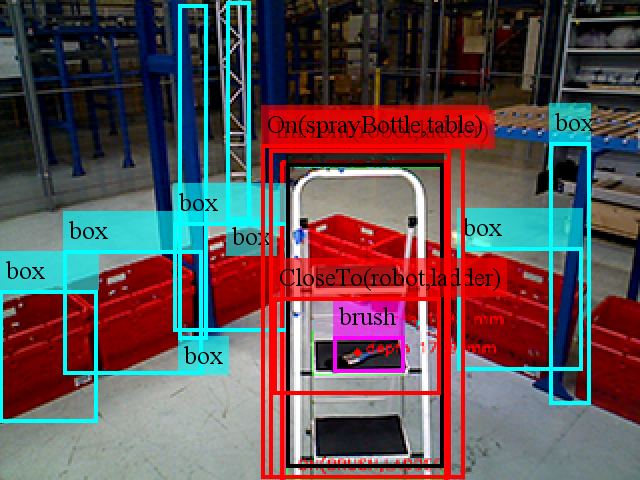}\hfill
    \caption[]{Recognition  during tasks execution.  The sequence shows the detection of \textit{guard}(panel), \textit{handle} and its manipulation to lower it helping the technician to hold the guard for inspecting the rollers.
The involved relations are $At$,$Hold$, $InFront$, $On$, and $CloseTo$. }
   \label{fig:tasks_sequences}
 \end{figure}

\paragraph{\em\bf Experiments of the VDEM framework at warehouse}

\begin{table}[h!]
\caption{Accuracy and average execution time according to task and configuration.}
\label{table:ablation}
\centering
\resizebox{.75\textwidth}{!}{
\begin{tabular}{|c|c|c|c|c|}
 \hline
 & & \textbf{\textit{PL + Ex + Kn}} & \textbf{\textit{PL + Ex + M}} & \textbf{\textit{PL + Ex + M + GPr}} \\ \hline
 &{Task 1} & {540 s} & {135 s}  &  {135 s}\\
 &{Task 2} & {260 s} & {70 s} & {70 s}   \\
{\textbf{a. ex. time}} & {Task 3}& {596 s} & {147 s}  & {147 s} \\
 & {Task 4}& {477 s} & {121 s} & {121 s}\\ 
 & {Task 5}& x & {52 s} & {52 s} \\ \hline
 
&{Task 1} & 23 & 72 & 81 \\
 &{Task 2}& 52 & 78 & 80\\
{\textbf{accuracy (\%)}} & {Task 3}& 24 & 68 & 79 \\
 & {Task 4}& 26 & 75 & 86\\ 
 & {Task 5}& x & 85 & 93 \\ \hline
\end{tabular}
}
\end{table}

In this section, we report the results of the experiments carried out with the VDEM deployed on the humanoid robot inside the warehouse. In the absence of other frameworks to make a comparison with, we perform a comprehensive ablation study. Table \ref{table:ablation} shows the results. 
 We identify the components of our framework with: \textit{PL} = Planning, \textit{Ex} = Execution, \textit{M} = Monitoring (Visual Stream), \textit{GPr} = Goals Prediction (LSTM). Furthermore, we indicate with \textit{Kn} the complete knowledge of the world.  

 The experiments were performed on 5 tasks: \textit{remove panel, support panel, clean diverter, bring object, find object}. Snapshots taken from two of these tasks are shown in Figure \ref{fig:tasks_sequences}. Each task was executed 50 times for assessing the accuracy, excluding failures caused by the robot controllistic part (grasping failure, platform movement error, etc). The tasks have been tested for each framework configuration, making 750 total experiments. Note that for Task 5, there are no values related to the first configuration. This is because  this task intrinsically requires perceptive and search skills, which can not be tested in the first configuration.
 
Starting from the \textit{PL + Ex  + Kn} case, the framework is tested with the FastDownward(FD) \cite{helmert-2006}  based planning system and the execution component. FD was adopted as it proved to be the fastest among the other planners that were considered, i.e. POMDP and PKS \cite{petrick-2004}. In this configuration a complete knowledge of the world was provided.
 We note that the system in this case suffers from long planning times caused by considering knowledge of the entire scene. Furthermore, this setting excludes dynamic and non-deterministic tasks.

Considering the \textit{PL + Ex + M} setting, the robot is able to complete all the tasks correctly, as it is possible to manage the non-deterministic nature of the tasks in this case. An example of the detection and monitoring capacity is shown in the first row of Figure \ref{fig:tasks_sequences}. 

 A limitation of this setting concerns the management of failures due to the inability  to predict the correct sequence of the goals.

Finally, the complete configuration of the framework is taken into consideration, \textit{PL + Ex + M + GPr}. In this setting tasks are decomposed and executed dynamically, identifying in real time different ways to complete a task. 
 A direct consequence of this greater flexibility, as can be seen in Table \ref{table:ablation}, is the improvement of the accuracy on the successful execution of the tasks. 

 An example is shown at the bottom row of Figure \ref{fig:tasks_sequences}. 
 In this case the task is to find, grab and bring the brush to the technician. Based on experience, the seq2seq system first suggests \textit{on(brush, table)}. 

 The goal fails, as another object is found (\textit{on(spraybottle, table)} detected). At this point the possibility of recovery using seq2seq comes into play. The execution monitor takes the second proposal (regarding the first goal to be achieved) made by the seq2seq-based proposal system, namely \textit{on(brush, ladder)}. 
 \vspace*{-3mm}

 \typeout{-------------------- Conclusion -----------------------}
\section{Conclusions}
We have presented an approach to vision based deep execution monitor for a robot assistive task. Both the idea and the realization are novel and promising. The experiment with the humanoid robot created at 
the High Performance Humanoid Technologies Lab (H2T) have proved that the framework proposed works as far as the specific tasks are considered and as far as the high level actions are taken into account. Weak elements of the approach are the ability of the robot to search the environment, which should cope with the limitation of vision at distances greater than 2.5 mt.  We are currently facing this problem by modeling search with deep reinforcement learning, so that the robot can optimize its search of objects and relations. 

\section{Acknowledgments}
The research has been granted by the H2020 Project SecondHands under grant agreement No 643950.
We thanks in particular our partners: the team at Ocado, Graham Deacon, Duncan Russel, Giuseppe Cotugno and Dario Turchi, the team of KIT led by Tamim Asfour, the team at UCL with Lourdes Agapito,  Martin Runz and Denis Tome, and the group at EPFL led by Aude Billiard.








\end{document}

%% file: ALGO.tex
\renewcommand{\vec}[1]{\mathbf{#1}}
\newcommand{\homog}[1]{\tilde{#1}}
\newcommand{\DM}{\mathrm{M}}
\newcommand{\imset}{\mathcal{J}}
\newcommand{\indset}{\mathcal{I}}
\newcommand{\pu}{u}
\newcommand{\pv}{v}

\begin{algorithm}%
\caption{Vision based deep execution monitor} \label{algo:vbdem}%
\KwIn{Current task ${\mathcal T}$, plan $\Pi$, current state $s$, plan library $Lib_{\Pi}$}%
\KwOut{end-task ${\mathcal T}$}%
\While{not end-task}{%
  \If{$\Pi\neq\emptyset$ and  $s = \bigwedge_i^N R_i({\boldsymbol \nu},a)$}{
    {$(\alpha,bounding\_box,depth) {:=} query{\_}vision(s)$}\\
    \If{$\alpha = True$}{
      \If{$s$ goal  of $\Pi$}{
        {$\Pi {:=} query{\_}seq2seq({\mathcal T},s,Lib_{\Pi})$}} 
        \Else{Continue $\Pi$}} 
     \Else{Return end-task ${\mathcal T}$}
 }
  \If{$\Pi = \emptyset$ and  $s = start({\mathcal T},s_0)$}{
     {$\Pi {:=} query{\_}seq2seq({\mathcal T},s,Lib_{\Pi})$}}
 \If{$\Pi\neq\emptyset$ and  $s = goal({\mathcal T})$}{
      {Return  $end$-$task {\mathcal T}$}}
    }
  \end{algorithm}%

\begin{algorithm}%
\caption{Query seq2seq} \label{algo:lstm}%
\KwIn{ seq2seq model, plan library $Lib_{\Pi}$, current task ${\mathcal T}$, current state $s$}%
\KwOut{subplan $\Pi$}%
            {Compute seq2seq output with input $({\mathcal T},s)$ and choose the goal state $s_g$ maximizing:
               $p(s_g | s,{\mathcal T})$}\\
             {Search  in $Lib_{\Pi}$  best match $\Pi$ with ${\boldsymbol \nu}, \{R\}_i^M$, mentioned in $s_g$, goal of $\Pi$}\\
           {Return $\Pi$}
\end{algorithm}%

\begin{algorithm}%
\caption{Query Vision} \label{algo:vision}%
\KwIn{video-stream at current time lapse $t_i{:}t_{i+n}$, DCNN models ${\mathcal M}_1,{\ldots},{\mathcal M}_k $,  current state $s$, thresholds $\mu,\tau$}%
\KwOut{Boolean}%
{$s = \bigwedge_i^N R_i({\boldsymbol\nu},a)$}\\
{Compute bounding boxes in video-stream using models ${\mathcal M}_1,{\ldots},{\mathcal M}_k $ }\\ 
{Segment objects in depth images in video-stream for each $\nu\in{\boldsymbol \nu}$   (Section \ref{sec:visualstream})}\\
\If{$confidence({\boldsymbol \nu})>\mu$ \ \ }{
  {Compute $R_i, i{=}1,{\ldots}N$ }}
\Else{
\While{time lapse $T< \tau$}{%
  {Search for missed $\nu\in{\boldsymbol \nu}$}}}
\If{$T\leq \tau$}{
  {Return $True$, bounding box for ${\boldsymbol \nu}$, depth}
}
\Else{Return ($False$,$\{\}$,$-1$)}
\end{algorithm}%

%% file: main.bbl
\begin{thebibliography}{10}
\providecommand{\url}[1]{#1}
\csname url@samestyle\endcsname
\providecommand{\newblock}{\relax}
\providecommand{\bibinfo}[2]{#2}
\providecommand{\BIBentrySTDinterwordspacing}{\spaceskip=0pt\relax}
\providecommand{\BIBentryALTinterwordstretchfactor}{4}
\providecommand{\BIBentryALTinterwordspacing}{\spaceskip=\fontdimen2\font plus
\BIBentryALTinterwordstretchfactor\fontdimen3\font minus
  \fontdimen4\font\relax}
\providecommand{\BIBforeignlanguage}[2]{{%
\expandafter\ifx\csname l@#1\endcsname\relax
\typeout{** WARNING: IEEEtranS.bst: No hyphenation pattern has been}%
\typeout{** loaded for the language `#1'. Using the pattern for}%
\typeout{** the default language instead.}%
\else
\language=\csname l@#1\endcsname
\fi
#2}}
\providecommand{\BIBdecl}{\relax}
\BIBdecl

\bibitem{Al-Omari-2016}
M.~Al-Omari, E.~Chinellato, Y.~Gatsoulis, D.~C. Hogg, and A.~G. Cohn,
  ``Unsupervised grounding of textual descriptions of object features and
  actions in video.'' in \emph{(KR),2016}, 2016, pp. 505--508.

\bibitem{alford2016}
R.~Alford, V.~Shivashankar, M.~Roberts, J.~Frank, and D.~W. Aha, ``Hierarchical
  planning: Relating task and goal decomposition with task sharing.'' in
  \emph{IJCAI}, 2016, pp. 3022--3029.

\bibitem{bahdanau2014neural}
D.~Bahdanau, K.~Cho, and Y.~Bengio, ``Neural machine translation by jointly
  learning to align and translate,'' \emph{arXiv preprint arXiv:1409.0473},
  2014.

\bibitem{bertsekas1995}
D.~P. Bertsekas and J.~N. Tsitsiklis, ``Neuro-dynamic programming: an
  overview,'' in \emph{Decision and Control, 1995.}, vol.~1, 1995, pp.
  560--564.

\bibitem{boutilier2000}
C.~Boutilier, R.~Reiter, M.~Soutchanski, S.~Thrun \emph{et~al.},
  ``Decision-theoretic, high-level agent programming in the situation
  calculus,'' in \emph{AAAI/IAAI}, 2000, pp. 355--362.

\bibitem{cohn2001qualitative}
A.~G. Cohn and S.~M. Hazarika, ``Qualitative spatial representation and
  reasoning: An overview,'' \emph{Fun. inf.}, vol.~46, no. 1-2, pp. 1--29,
  2001.

\bibitem{damen2018}
D.~Damen, H.~Doughty, G.~M. Farinella, S.~Fidler, A.~Furnari, E.~Kazakos,
  D.~Moltisanti, J.~Munro, T.~Perrett, W.~Price \emph{et~al.}, ``Scaling
  egocentric vision: The epic-kitchens dataset,'' in \emph{ECCV}, 2018.

\bibitem{das2016}
A.~Das, H.~Agrawal, C.~L. Zitnick, D.~Parikh, and D.~Batra, ``Human attention
  in visual question answering: Do humans and deep networks look at the same
  regions?'' \emph{arXiv preprint arXiv:1606.03556}, 2016.

\bibitem{doyle1986}
R.~J. Doyle, D.~J. Atkinson, and R.~S. Doshi, ``Generating perception requests
  and expectations to verify the execution of plans,'' in \emph{AAAI}, 1986,
  pp. 81--88.

\bibitem{erol1994}
K.~Erol, J.~A. Hendler, and D.~S. Nau, ``Umcp: A sound and complete procedure
  for hierarchical task-network planning.'' in \emph{AIPS}, vol.~94, 1994, pp.
  249--254.

\bibitem{Everingham15}
M.~Everingham, S.~M.~A. Eslami, L.~Van~Gool, C.~K.~I. Williams, J.~Winn, and
  A.~Zisserman, ``The pascal visual object classes challenge: A
  retrospective,'' \emph{IJCV}, vol. 111, no.~1, pp. 98--136, 2015.

\bibitem{fikes1971}
R.~E. Fikes, ``Monitored execution of robot plans produced by strips,'' SRI,
  Tech. Rep., 1971.

\bibitem{finzi2001}
A.~Finzi and F.~Pirri, ``Combining probabilities, failures and safety in robot
  control,'' in \emph{INTERNATIONAL JOINT CONFERENCE ON ARTIFICIAL
  INTELLIGENCE}, vol.~17, no.~1.\hskip 1em plus 0.5em minus 0.4em\relax
  LAWRENCE ERLBAUM ASSOCIATES LTD, 2001, pp. 1331--1336.

\bibitem{furnari2017next}
A.~Furnari, S.~Battiato, K.~Grauman, and G.~M. Farinella, ``Next-active-object
  prediction from egocentric videos,'' \emph{Journal of Visual Communication
  and Image Representation}, vol.~49, pp. 401--411, 2017.

\bibitem{gianni2015}
M.~Gianni, G.-J.~M. Kruijff, and F.~Pirri, ``A stimulus-response framework for
  robot control,'' \emph{ACM Transactions on Interactive Intelligent Systems
  (TiiS)}, vol.~4, no.~4, p.~21, 2015.

\bibitem{guadarrama2013}
S.~Guadarrama, L.~Riano, D.~Golland, D.~Go, Y.~Jia, D.~Klein, P.~Abbeel,
  T.~Darrell \emph{et~al.}, ``Grounding spatial relations for human-robot
  interaction,'' in \emph{IROS}, 2013, pp. 1640--1647.

\bibitem{gupta2017}
S.~Gupta, J.~Davidson, S.~Levine, R.~Sukthankar, and J.~Malik, ``Cognitive
  mapping and planning for visual navigation,'' \emph{arXiv preprint
  arXiv:1702.03920}, vol.~3, 2017.

\bibitem{haarnoja2016}
T.~Haarnoja, A.~Ajay, S.~Levine, and P.~Abbeel, ``Backprop kf: Learning
  discriminative deterministic state estimators,'' in \emph{Adv. in Neural Inf.
  Proc. Sys.}, 2016, pp. 4376--4384.

\bibitem{he2017}
K.~He, G.~Gkioxari, P.~Doll{\'a}r, and R.~Girshick, ``Mask r-cnn,'' in
  \emph{Computer Vision (ICCV), 2017 IEEE International Conference on}.\hskip
  1em plus 0.5em minus 0.4em\relax IEEE, 2017, pp. 2980--2988.

\bibitem{helmert-2006}
M.~Helmert, ``The fast downward planning system.'' \emph{JAIR}, vol.~26, pp.
  191--246, 2006.

\bibitem{hochreiter1997}
S.~Hochreiter and J.~Schmidhuber, ``Long short-term memory,'' \emph{Neural
  computation}, vol.~9, no.~8, pp. 1735--1780, 1997.

\bibitem{hofmann2017}
T.~Hofmann, T.~Niemueller, and G.~Lakemeyer, ``Initial results on generating
  macro actions from a plan database for planning on autonomous mobile
  robots,'' 2017.

\bibitem{hornung2014}
A.~Hornung, S.~B{\"o}ttcher, J.~Schlagenhauf, C.~Dornhege, A.~Hertle, and
  M.~Bennewitz, ``Mobile manipulation in cluttered environments with humanoids:
  Integrated perception, task planning, and action execution,'' in
  \emph{Humanoids}, 2014, pp. 773--778.

\bibitem{ingrand2017}
F.~Ingrand and M.~Ghallab, ``Deliberation for autonomous robots: A survey,''
  \emph{Artificial Intelligence}, vol. 247, pp. 10--44, 2017.

\bibitem{karkus2017}
P.~Karkus, D.~Hsu, and W.~S. Lee, ``Qmdp-net: Deep learning for planning under
  partial observability,'' in \emph{Adv. in Neural Inf. Proc. Sys.}, 2017, pp.
  4697--4707.

\bibitem{krizhevsky2012}
A.~Krizhevsky, I.~Sutskever, and G.~E. Hinton, ``Imagenet classification with
  deep convolutional neural networks,'' in \emph{NIPS}, 2012, pp. 1097--1105.

\bibitem{lenz2015}
I.~Lenz, H.~Lee, and A.~Saxena, ``Deep learning for detecting robotic grasps,''
  \emph{Int J. of Robotics Res.}, vol.~34, no. 4-5, pp. 705--724, 2015.

\bibitem{littman2002}
M.~L. Littman and R.~S. Sutton, ``Predictive representations of state,'' in
  \emph{Adv. in neural inf. proc. sys.}, 2002, pp. 1555--1561.

\bibitem{Lu-2016}
C.~Lu, R.~Krishna, M.~Bernstein, and L.~Fei-Fei, ``Visual relationship
  detection with language priors,'' in \emph{{ECCV}}, 2016, pp. 852--869.

\bibitem{luong2015a}
M.-T. Luong, H.~Pham, and C.~D. Manning, ``Effective approaches to
  attention-based neural machine translation,'' \emph{arXiv preprint
  arXiv:1508.04025}, 2015.

\bibitem{luong2014}
M.-T. Luong, I.~Sutskever, Q.~V. Le, O.~Vinyals, and W.~Zaremba, ``Addressing
  the rare word problem in neural machine translation,'' \emph{arXiv preprint
  arXiv:1410.8206}, 2014.

\bibitem{mendoza2015}
J.~P. Mendoza, M.~Veloso, and R.~Simmons, ``Plan execution monitoring through
  detection of unmet expectations about action outcomes,'' in \emph{ICRA},
  2015, pp. 3247--3252.

\bibitem{mirowski2016}
P.~Mirowski, R.~Pascanu, F.~Viola, H.~Soyer, A.~Ballard, A.~Banino, M.~Denil,
  R.~Goroshin, L.~Sifre, K.~Kavukcuoglu \emph{et~al.}, ``Learning to navigate
  in complex environments,'' \emph{arXiv:1611.03673}, 2016.

\bibitem{nilsson1973}
N.~J. Nilsson, \emph{A hierarchical robot planning and execution system}.\hskip
  1em plus 0.5em minus 0.4em\relax SRI, 1973.

\bibitem{ntouskos2013}
V.~Ntouskos, F.~Pirri, M.~Pizzoli, A.~Sinha, and B.~Cafaro, ``Saliency
  prediction in the coherence theory of attention,'' \emph{Biologically
  Inspired Cognitive Architectures}, vol.~5, pp. 10--28, 2013.

\bibitem{ntouskos2015}
V.~Ntouskos, M.~Sanzari, B.~Cafaro, F.~Nardi, F.~Natola, F.~Pirri, and M.~Ruiz,
  ``Component-wise modeling of articulated objects,'' in \emph{Proceedings of
  the IEEE International Conference on Computer Vision}, 2015, pp. 2327--2335.

\bibitem{petrick-2004}
R.~P. Petrick and F.~Bacchus, ``Pks: Knowledge-based planning with incomplete
  information and sensing,'' \emph{Proc. of the Sys. Dem. sess. at ICAPS},
  2004.

\bibitem{pettersson2005}
O.~Pettersson, ``Execution monitoring in robotics: A survey,'' \emph{Robotics
  and Autonomous Systems}, vol.~53, no.~2, pp. 73--88, 2005.

\bibitem{raffel2017}
C.~Raffel, M.-T. Luong, P.~J. Liu, R.~J. Weiss, and D.~Eck, ``Online and
  linear-time attention by enforcing monotonic alignments,'' \emph{arXiv
  preprint arXiv:1704.00784}, 2017.

\bibitem{ren-2015}
S.~Ren, K.~He, R.~Girshick, and J.~Sun, ``Faster r-cnn: Towards real-time
  object detection with region proposal networks,'' in \emph{NIPS}, 2015, pp.
  91--99.

\bibitem{sabharwal2011}
C.~L. Sabharwal, J.~L. Leopold, and N.~Eloe, ``A more expressive 3d region
  connection calculus.'' in \emph{DMS}.\hskip 1em plus 0.5em minus 0.4em\relax
  Citeseer, 2011, pp. 307--311.

\bibitem{santoro2017}
A.~Santoro, D.~Raposo, D.~G. Barrett, M.~Malinowski, R.~Pascanu, P.~Battaglia,
  and T.~Lillicrap, ``A simple neural network module for relational
  reasoning,'' in \emph{Adv. in neural inf. proc. sys.}, 2017, pp. 4974--4983.

\bibitem{sanzari2016}
M.~Sanzari, V.~Ntouskos, and F.~Pirri, ``Bayesian image based 3d pose
  estimation,'' in \emph{ECCV 2016}, 2016, pp. 566--582.

\bibitem{shivashankar2015}
V.~Shivashankar, ``Hierarchical goal networks: Formalisms and algorithms for
  planning and acting,'' Ph.D. dissertation, University of Maryland, College
  Park, 2015.

\bibitem{simonyan2014}
K.~Simonyan and A.~Zisserman, ``Very deep convolutional networks for
  large-scale image recognition,'' \emph{arXiv:1409.1556}, 2014.

\bibitem{sutskever2014sequence}
I.~Sutskever, O.~Vinyals, and Q.~V. Le, ``Sequence to sequence learning with
  neural networks,'' in \emph{Advances in neural information processing
  systems}, 2014, pp. 3104--3112.

\bibitem{sutskever2014}
------, ``Sequence to sequence learning with neural networks,'' in \emph{Adv.
  in neural inf. proc. sys.}, 2014, pp. 3104--3112.

\bibitem{sutton1998}
R.~S. Sutton and A.~G. Barto, \emph{Reinforcement learning: An introduction
  (Second Edition)}, 1998,2017, vol.~1, no.~1.

\bibitem{wachter2016}
M.~W{\"a}chter, S.~Ottenhaus, M.~Kr{\"o}hnert, N.~Vahrenkamp, and T.~Asfour,
  ``The armarx statechart concept: graphical programing of robot behavior,''
  \emph{Frontiers in Robotics and AI}, vol.~3, p.~33, 2016.

\bibitem{wang2017}
H.~Wang, W.~Liang, and L.-F. Yu, ``Transferring objects: Joint inference of
  container and human pose,'' in \emph{CVPR}, 2017, pp. 2933--2941.

\bibitem{wilkins1985}
D.~E. Wilkins, ``Recovering from execution errors in sipe,''
  \emph{Computational. Intelligence}, vol.~1, no.~1, pp. 33--45, 1985.

\bibitem{wu2017}
C.~Wu, J.~Zhang, O.~Sener, B.~Selman, S.~Savarese, and A.~Saxena,
  ``Watch-n-patch: unsupervised learning of actions and relations,''
  \emph{TPAMI}, 2017.

\bibitem{zeiler2014}
M.~D. Zeiler and R.~Fergus, ``Visualizing and understanding convolutional
  networks,'' in \emph{ECCV}, 2014, pp. 818--833.

\bibitem{zhu2017}
L.~Zhu, Z.~Xu, Y.~Yang, and A.~G. Hauptmann, ``Uncovering the temporal context
  for video question answering,'' \emph{IJCV}, vol. 124, no.~3, pp. 409--421,
  2017.

\bibitem{ZhuGKFFGMF17}
Y.~Zhu, D.~Gordon, E.~Kolve, D.~Fox, L.~Fei{-}Fei, A.~Gupta, R.~Mottaghi, and
  A.~Farhadi, ``Visual semantic planning using deep successor
  representations,'' \emph{CoRR}, 2017.

\end{thebibliography}
